\documentclass{article}

\usepackage{arxiv}

\usepackage[utf8]{inputenc} 
\usepackage[T1]{fontenc}    
\usepackage{hyperref}       
\usepackage{url}            
\usepackage{booktabs}       
\usepackage{amsfonts}       
\usepackage{nicefrac}       
\usepackage{microtype}      
\usepackage{lipsum}		
\usepackage{graphicx}
\usepackage{natbib}
\usepackage{doi}
 \usepackage{float}

\title{Relating CNNs with brain: Challenges and findings}

\author{ \href{https://orcid.org/0000-0002-4594-211X}{\includegraphics[scale=0.06]{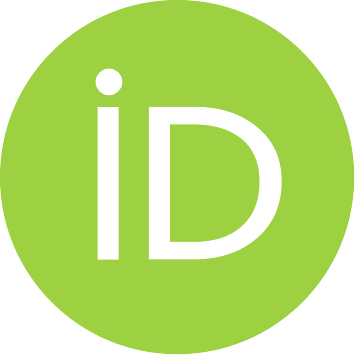}\hspace{1mm}Reem Abdel-Salam}\\
	Department of Computer engineering\\
	Cairo University\\

	\texttt{reem.abdelsalam13@gmail.com} \\

}

\renewcommand{\shorttitle}{\textit{arXiv} Template}

\hypersetup{
pdftitle={A template for the arxiv style},
pdfsubject={q-bio.NC, q-bio.QM},
pdfauthor={David S.~Hippocampus, Elias D.~Striatum},
pdfkeywords={First keyword, Second keyword, More},
}

\begin{document}
\maketitle

\begin{abstract}
Conventional neural network models (CNN), loosely inspired by the primate visual system, have been shown to predict neural responses in the visual cortex. However, the relationship between CNNs and the visual system is incomplete due to many reasons. On one hand state of the art CNN architecture is very complex, yet can be fooled by imperceptibly small, explicitly crafted perturbations which makes it hard difficult to map layers of the network with the visual system and to understand what they are doing. On the other hand, we don't know the exact mapping between feature space of the CNNs and the space domain of the visual cortex, which makes it hard to accurately predict neural responses. In this paper we review the challenges and the methods that have been used to predict neural responses in the visual cortex and whole brain as part of The Algonauts Project 2021 Challenge:
``How the Human Brain Makes Sense of a World in Motion". \citep{cichy2021algonauts}.

\end{abstract}


\section{Introduction}
Building visual encoding models that accurately predict visual responses is a major challenge for today's vision-based brain-machine interface techniques. This goal can be met by employing functional magnetic resonance imaging (fMRI), which measures brain activity in human subjects who are passively viewing images/videos \citep{kay2018principles}. Visual encoding models, in particular, can be built based on blood-oxygen-level-dependent (BOLD) signals to predict neural responses to arbitrary stimuli and thus develop a better understanding of visual information processing in the human brain. The visual processing in the brain from external visual stimuli is usually non-linear. In order to simulate non-linear mapping between input visual stimuli and brain activity, visual encoding models are used. However, one significant obstacle in this modeling and prediction process is the difficulty in developing an effective non-linear feature space to study the non-linearity of neural responses \citep{naselaris2011encoding}.\\
In early visual encoding models, feature space are mainly constructed using hand-crafted features inspired by some visual representation mechanisms. \cite{kay2008identifying} used the Gabor wavelet pyramid model as the nonlinear feature extractor from the stimulus space to the feature space, and receptive-field models to predict brain activity. The Gabor wavelet pyramid became a classical encoding model for low-level visual information, especially when predicting brain activity in the V1 region. Many following studies enhanced the Gabor wavelet pyramid model's fundamental characteristics and prediction outputs as in \citep{huth2012continuous,stansbury2013natural}.
\cite{nishimoto2011reconstructing} proposed using a fixed set of nonlinear spatiotemporal motion-energy filters and then a set of hemodynamic response filters fit separately to each voxel. Encoding models based on handcrafted features have high biological interpretability but poor prediction performance and universality because they are often created for a few specific visual regions.\\
Deep neural networks (DNNs) with hierarchical feature transformation have recently made achievements in a range of disciplines (including computer vision), capturing the attention of computational neuroscience researchers. Numerous neuroscience methods have used convolutional neural networks (CNNs) to represent the human visual system, achieving tremendous advancements in building visual encoding models. In general, these CNN-based visual encoding modeling approaches can be divided into task-driven and data-driven approaches. In task-driven approaches usually, neural responses are generated by using activation features in intermediate layers in pre-trained CNNs on a vision task as imagenet. \cite{agrawal2014pixels} investigated two models  Fisher Vectors and Convnets to produce features in order to predict brain activity across many low- and high-level visual
areas. \cite{gucclu2015deep} used two different CNNs for layer-wise analysis of voxel scores in the ventral stream.
\cite{shi2018deep} proposed adding a recurrent neural network (RNN) to a pre-trained CNN (VGG16) in order to predict cortical fMRI responses to natural movie stimuli. The RNN better predicted cortical responses to natural movie stimuli
than the CNN, at all visual areas, especially those along the dorsal stream. This demonstrates the potential of using the RNN for in-depth computational understanding of the dynamic natural vision. \cite{han2019variational} proposed a variational auto-encoder (VAE), as a computational model of the visual cortex. The first five layers of the VAE acts as an encoder while the last five layers act as a decoder to learn visual representations from a diverse set of unlabeled images.
To elicit visual cortical responses, data-driven techniques immediately train all parameters of a given encoding model based on experimental inputs. \cite{zhang2019visual} used transfer learning to incorporate a previously trained DNN and build a nonlinear mapping from visual features to brain activity. This nonlinear mapping replaces traditional linear mapping and is intended to improve brain activity prediction accuracy. \cite{qiao2021effective} proposed an end-to-end CNN regression model for visual encoding based on fMRI data. As a result, this model could develop appropriate feature representations and linear regression weights for visual cortical responses while also significantly improving prediction performance. \cite{cui2021gabornet} proposed GaborNet visual encoding an end to end encoding model, combining handcrafted and deep ROI features. The key idea behind it is to have the kind of preferred hand-crafted features in V1 and V2 areas. \\
The Algonauts Project 2021 Challenge:
How the Human Brain Makes Sense of a World in Motion \citep{cichy2021algonauts} focuses on event understanding from videos, mainly to predict human brain responses for short video clips of everyday events (e.g., panda eating, fish swimming, a
person paddling). The Algonauts Project 2021 Challenge had two tracks mini-track where the goal is to predict brain activity in 9 ROIs of the visual brain (V1, V2, V3, V4, Body- EBA, Face - FFA, STS, Object - LOC, Scene - PPA), the full-track goal is to predict the whole-brain neural activity. \\
In this paper, we discuss the work performed on both tracks. The rest of this paper is structured as follows: section \ref{data_used} describes the dataset used in both tracks: mini-track and full track, section \ref{methedology} discusses the approaches used in our study, and section \ref{results} discusses the obtained results and analysis.

\section{Dataset}
\label{data_used}
The dataset provided contains 10102 3- seconds videos. The first 1000 videos are considered as training set and are associated with fMRI human brain data of 10 subjects in response to viewing muted videos from this set. The test set is the last 102 3 seconds videos. For the mini-track, The data is provided for 9 regions of the visual brain (V1, V2, V3, V4, LOC, EBA, FFA, STS, PPA) in a Pickle file (e.g. V1.pkl) that contains a num\_videos x num\_repetitions x num\_voxels matrix. For each ROI, while for the full track The data is provided for selected voxels across the whole brain in a Pickle file (e.g. WB.pkl) that contains a num\_videos x num\_repetitions x num\_voxels matrix.

\section{Materials and Methods}
\label{methedology}
The aim of the challenge is to predict neural activity (fMRI responses) in the visual cortex for 9 ROIs of the visual brain (V1, V2, V3, V4, Body- EBA, Face - FFA, STS, Object - LOC, Scene - PPA)  in the mini track challenge and to predict the whole-brain neural activity (fMRI responses) in the full track challenge. The most known approaches to predict fMRI responses are Voxel-wise encoding models, ROI-wise encoding models. In Voxel-wise encoding model \citep{wu2006complete,naselaris2011encoding}, features are either hand-crafted (Gabor Wavelet for V1,...etc) or extracted from well known CNNs (Pixels to feature space). Then these features are either kept as is, or their dimensionality is reduced. Those features are thought of as potential hypothesis features that are generated or similar to the one generated in the brain region, represented as a stimulus. Afterward, these features are linearly mapped onto each voxel’s responses using a suitable computational model (linear regression, multi-layer perceptron,...etc)
Unlike the Voxel-wise encoding model where thousands of regression
models are constructed for several visual regions of interest (ROIs), in ROI-wise encoding models, all voxels in one region of interest (ROI) are jointly encoded.
Two types of experiments for Stimulus-model-based-encoding were conducted. The first approach is transfer learning of the deep neural network to incorporate a pre-trained DNN  and
train a nonlinear mapping from visual features to brain activity similar to what was discussed in \citep{zhang2019visual}. The second approach is following the traditional pipeline for Voxel-wise encoding models.
\subsection{Transfer learning for Stimulus prediction}
The aim of the challenge is to predict neural activity responses to a set of 3-second videos, for the mini-challenge in the 9 ROIs or whole brain for the full challenge. This experiment was conducted for the mini-challenge only. Usually, for this type of experiment, the input is usually an image and its corresponding stimulus. Since the input in this challenge is 3 seconds videos, the challenge was how to perform the same task using non-video CNNs (as they were hard to load and train using colab, without having notebook to crash even for a small batch size). The aim was to build/finetune a network that generates the stimulus for a given video.

The general process for transfer learning goes as follows:
\begin{itemize}
  \item A CNN  network is trained on specific tasks and data set. 
  \item This CNN (base model) is chosen which is the state of the art in this specific domain, or a model which has a near accuracy to the state of the art.
  \item N-layers from this chosen model are copied to a target network (new network with additional layers), in a new domain.
  \item Train the new model in the new domain, while keeping weights of the first N-layers fixed (i.e frozen) or finetune them.
\end{itemize}
The building process of our pipeline goes as follows:
\begin{itemize}
\item AlextNet, and resnet-18 semi-supervised  \citep{yalniz2019billion} models were chosen as our candidate base models.
\item For the AlextNet, the first  five convolutional layers, were copied  (with 96, 256, 384, 384, and 256
kernels, respectively), while leaving the three fully connected layers. For the resnet-18 semi-supervised model, all layers were kept except for the fully connected layer which was then removed.

\item For the target network three variants were introduced:
\begin{itemize}
\item Adding three fully connected layers after the base model.
  For AlextNet the three fully connected layers were of size 4096, 4096, and number of stimuli for the voxel in given region neurons. For the  resnet-18 semi-supervised model the three fully connected layers were of size 25088, 1024, and number of stimuli for the voxel in given region neurons.
  
\item Adding Two LSTM layers followed by two fully connected layers. For the AlextNet model, the two LSTM layers were bidirectional, with a hidden state of 512, and dropout of 0.2. The two fully connected layers of sizes 1024, number of stimulus for the voxel in given region neurons.

\item Adding An three LSTM layers followed by two fully connected layers. The idea behind it is that the stimulus is BOLD time series, therefore it might be better to generate time series data directly from the features. For the AlextNet model, the two LSTM layers were bidirectional, with a hidden state of 512, and dropout of 0.2. The last LSTM layer had a hidden state of the number of stimuli for the voxel in given region neurons.

\end{itemize}

\item The input video is segmented into 16 frames, for each video, The 16 frames images are inputted as a batch, and we have taken the mean values of their prediction as the predicted stimulus for a given ROI.
\item The prediction dissimilarity between the predicted stimulus and the original stimulus was used as a loss function for the network.
\item The layers of the base network were frozen, while we trained the additional added layers in the target network.
\item the target model was trained for 10 epochs, we kept the best model version, which had high correlation prediction in the validation set. We used 90-10 percent scheme as our training, validation policy.
\item Adam optimizer with learning rate 0.001 was used for training.

\end{itemize}
Based on a training dataset, voxel-wise encoding models can find the optimum model for each voxel. However, because it is computationally impossible to train a complicated non-linear model independently for each voxel, voxel-wise modeling can only use a linear mapping between the feature and the brain activity spaces. We used an ROI-wise encoding modeling strategy here, which means we trained one model to predict the responses of all voxels in each ROI. This method considerably decreases the training time and, more crucially, the model parameters are jointly restricted by the prediction accuracy across all voxels. There are multiple caveats in this system:  training process is more vulnerable to some noisy voxels, removing fully connected layers in the base models, where fully connect layers in the base models usually are good features for  EBA, FFA, STS,  LOC, PPA ROIs, and the optimal number of frames per video. In early experiments, MSE loss in addition to dissimilarity metric was used together, but they didn't seem to perform well, on the training and validation datasets. 

\subsection{Traditional pipeline for Voxel-wise encoding models}
In this experiment several deep CNN models for features extraction were used, in addition to that several computational model for generating voxels, were tested for both challenges.
VGG16, VGG19, resnet-18, resnet-18 semi-supervised, resnet-18 semi weakly supervised, resnet-50 semi-supervised, Nasnet mobile, AlexNet, Squeezenet, VOneNet resnset50 \citep{dapello2020simulating}, and VOneNet Alexnet were used as our feature extractor. For the computational models,
Elastic net, lasso regression, linear regression, ridge regression, and RES \citep{anderson2016representational} were used.
Our pipeline goes as follows:

\begin{itemize}
  \item Each video in the training and test set was converted into batches of image frames of size 15,10,16,26 per video.
  \item For each batch, different layer activations for a given CNN model is extracted. 
  \item These extracted features are then reduced in dimension using principal component analysis (PCA) with 100 feature (100 feature was optimal number for several CNNs and layers in training and validation set). Normalization is then applied to the dataset.
  \item For each ROI, the best feature (extracted from activation layer after applying PCA and normalization) with high correlation in the validation dataset for several subjects, is used for the training.
  
  \item For choosing the best computational model, The best model with a given feature and high correlation with respect to the validation set was chosen while applying cross-validation of 3.
  \item 90-10 train-validation split scheme was used.
  
\end{itemize}
Multiple variants from this pipeline were introduced, where we tried to combine 2 layers or more as features, generating a new feature matrix consisting of all polynomial combinations of the features with a higher degree, not applying PCA. However, the successful one was the pipeline discussed above. 
As stated in our pipeline above: activation layer features are either extracted from all CNN model layers or for a specific layers, for each training and testing examples. Then PCA and normalization are applied. We train multiple models for each subject and for each layer, for a specific ROI. Then the best perform modeling for a specific layer for all subjects is chosen for generating synthesized stimulus for the test set. 
For AlexNet model (which consists of 8 layers, 5 convolutional layers, and three fully connected layers) all layers were chosen for generating the features. For the ROIs, layer 5 was used for whole-brain prediction, layer 2 was used for V1 prediction. For V2, V3, V4 layer 5 was used for prediction. For LOC, EBA, FFA, PPA layer 8 was used, while for STS layer 6 is used.
For resnet-18, resnet-18 semi-supervised, and resnet18- semi weakly supervised models we used the first maxpool layer, four main blocks, and the fully connected layer as our feature set. For the ROIs, the last fully connected layer was used for whole-brain prediction, the third block was used for V1, V2, V3 regions. For V4 fourth block was used, while for the remaining ROIs the last fully connected layer was used. For resnet-50 and VOnet resnet-50 the last 15 Bottleneck layers were used in addition to the last fully connected layer. For the ROIs, the last Bottleneck was used for whole-brain prediction, in addition to V4, LOC, EBA, FFA, PPA. For V1-V3 10th bottleneck layer was used, while for STS the 14th bottleneck layer was used. For Nasnet mobile we used most of the layers for generating features, resulting in 18 different features. For the ROIs, layer 4 was used for WB, V2, V3, V4, LOC, FFA, STS prediction. For V1 prediction layer 2 was used, while layer 8 was used for PPA prediction. Usually, the first couple of layers are used for the prediction of the V1-V4 area, while later layers in the model are used for prediction LOC, FFA, PPA, STS, WB, and the pattern holds for most of the tested models.

\section{Evaluation and Results}
\label{results}

\subsection{Evaluation}
Evaluation for the test set on the leaderboard is carried out by comparing predict brain responses to the empirically measured brain responses. The comparison is carried out using Pearson’s correlation, comparing for each voxel the 102-dimensional vector formed by the activations for the 102 test set video clips. Then the correlation is normalized by the square root of the corresponding voxel’s split-half reliability. These results in one value of normalized correlation per voxel ranging from -1 to 1. Then the predictivity is averaged across selected voxels in the brain full track, or across all the voxels in each of the 9 brain regions of interest. For the model`s evaluation in the training and validation phase,  Pearson’s correlation was used for either a selected subject or multiple subjects for each ROI.
\subsection{Results}
\begin{figure*}[h]
  \centering
  \includegraphics[width=\linewidth,height=4in,width=\linewidth]{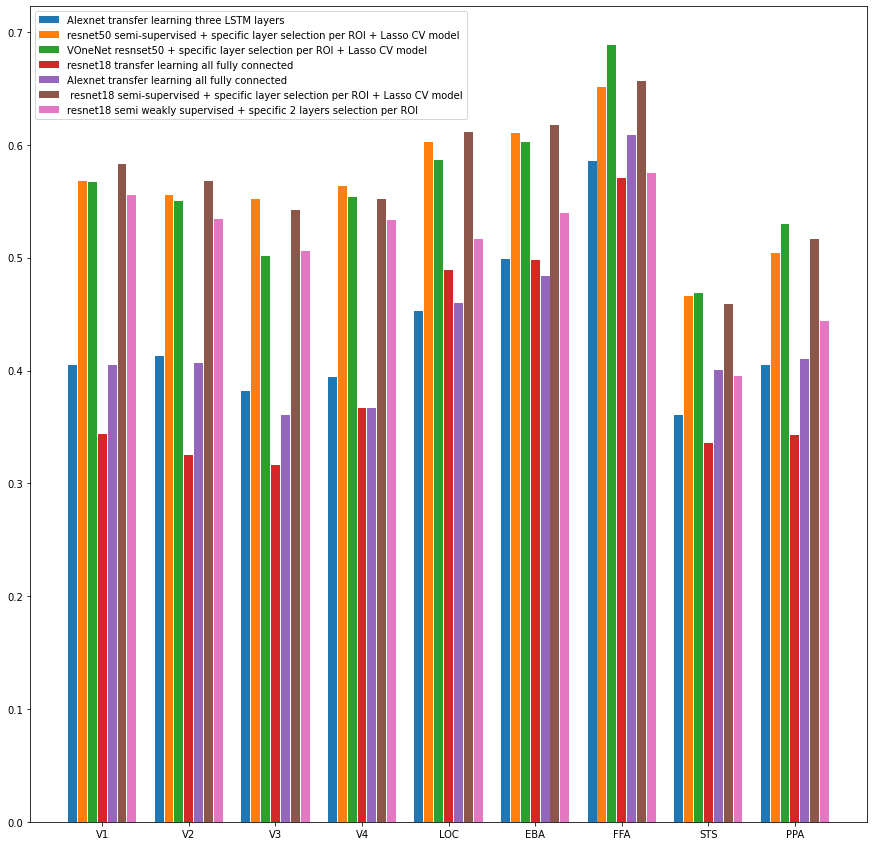}
  \caption{Comparison between different model with respect to the 9 ROIs in the mini track in the test set}
  \label{fig:ROIS test set}
\end{figure*}

\begin{figure*}[h]
  \centering
  \includegraphics[width=\linewidth,height=3in,width=\linewidth]{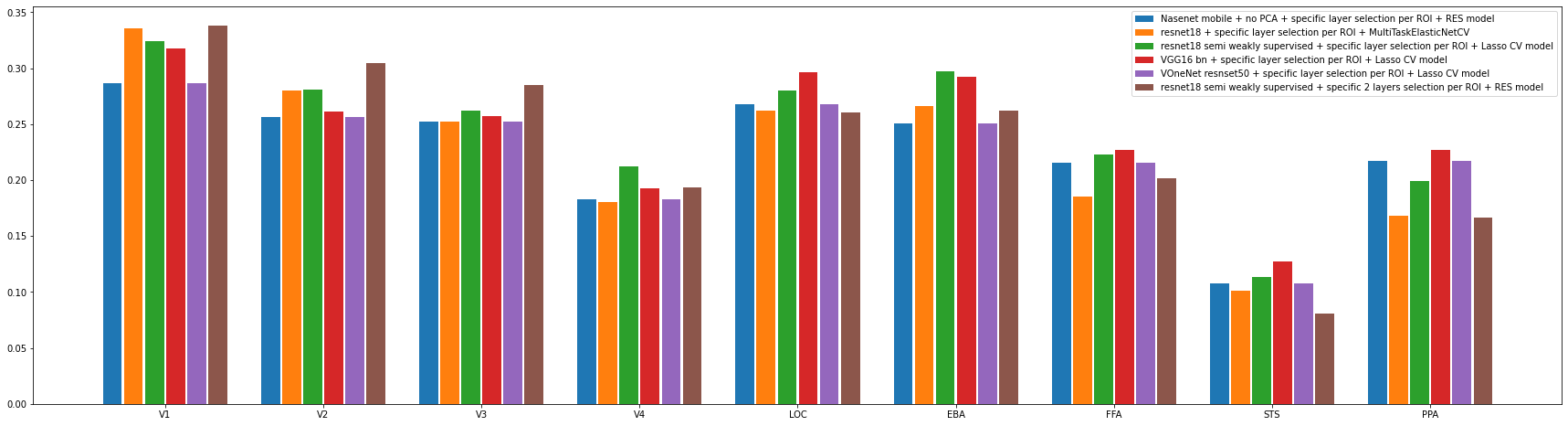}
  \caption{Comparison between different model with respect to the 9 ROIs in the mini track in the validation set for subject 4}
  \label{fig:ROIS validation set}
\end{figure*}

\begin{figure*}[h]
  \centering
  \includegraphics[width=\linewidth,height=3in,width=\linewidth]{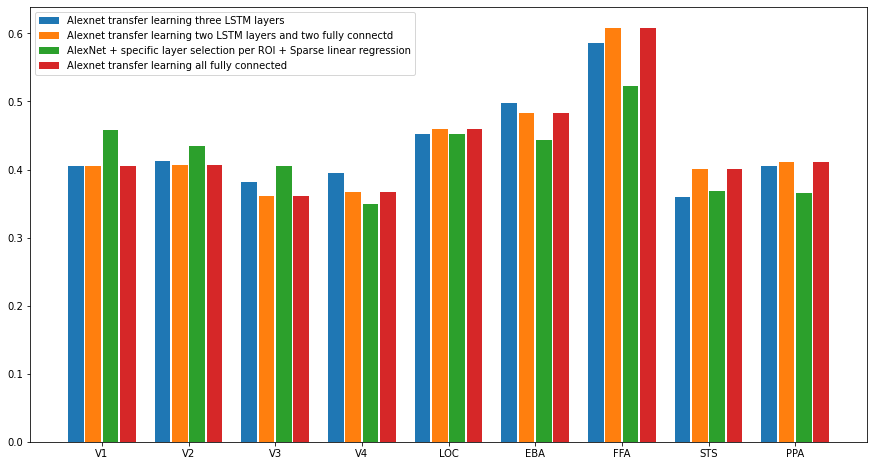}
  \caption{Comparison between different approaches used in Alexnet model}
  \label{fig:Alex net finetune comparison}
\end{figure*}

Table \ref{results_minitrack} shows the achieved score for each model in the mini track, while table  \ref{results_fulltrack} shows the achieved score for each model in the full track. The best performing model in the mini track was resnet-18 semi-supervised with lasso as a regression model, while for the full track Nasnet mobile model with RES as a model. Further analysis is conducted for the mini track results as shown in figure \ref{fig:ROIS test set}, resnet18 semi-supervised with specific layer selection per ROI and Lasso CV model has a higher correlation in V1, V2, Loc, and EBA regions, while resnet50 semi-supervised with specific layer selection per ROI and Lasso CV model performs well by having high correlation in V4, V3 areas. For FFA, STS, and PPA areas, VOneNet resnset50 with specific layer selection per ROI and Lasso CV model show a high correlation between the predicted stimulus and the actual one. In the early stages in the development part, we have chosen the model and the layer for each region, based on the one that shows a high correlation. As shown in figure \ref{fig:ROIS validation set}, there are several models that show better performance in the validation set, however badly performs on the test set. Therefore, it could be concluded that we cannot rely solely on the validation set to choose candidate models and layers. Different approaches on Alexnet model were analyzed on the test set, as shown in figure \ref{fig:Alex net finetune comparison}. AlexNet with specific layer selection per ROI and Sparse linear regression shows better correlation in V1, V2, and V3 areas, while for V4and EBA areas Alexnet transfer learning three LSTM layers shows better results. For LOC, FFA, STS, and PPA areas, the Alexnet transfer learning all fully connected shows a better correlation. The interpretability and accuracy of an encoding model are two important indicators to consider when evaluating it. The accuracy of a visual encoding model indicates how well it can predict the brain activity elicited by novel stimuli. The interpretability metric indicates how well we can capture the relationship between the encoding model's components (i.e., visual features) and the outcomes predicted by the model. Voxel-wise encoding models show higher prediction accuracy and correlation in all brain ROIs. However, it's not clear when ROI-wise encoding models with non-linear mapping outperforms voxel-wise encoding models. Most of the semi-supervised models and the weakly supervised models show higher performance in most ROIs, which suggests to uses such models features.

\begin{table}[h]
\begin{center}
\begin{tabular}{||c | c||} 
 \hline
 Method & Score\\ [0.5ex] 
 \hline\hline
 Alexnet + all layers as features + Sparse linear regression & 0.181348463  \\ 
 \hline
 AlexNet + specific layer selection per ROI + Sparse linear regression & 0.4222287033  \\
 \hline
 resnet18 + specific layer selection per ROI + Sparse linear regression & 0.4554744991  \\
 \hline
 resnet18 + specific layer selection per ROI + MultiTaskElasticNetCV & 0.503842829 \\
 \hline
 Alexnet transfer learning all fully connected & 0.4318425715 \\ \hline
 Alexnet transfer learning three LSTM layers & 0.4328641541 \\ \hline
 resnet18 transfer learning all fully connected & 0.3986074061	 \\ \hline
 
 Alexnet transfer learning two LSTM layers and two fully connected layers & 0.433419799 \\ \hline
 
 resnet18 + specific layer selection per ROI +RES model & 0.5000028286	 \\\hline 
 Nasenet mobile + no PCA + specific layer selection per ROI + RES model & 0.4999065559	 \\\hline
 VGG16 bn + specific layer selection per ROI + Lasso CV model & 0.0578059194		 \\\hline
 resnet18 semi weakly supervised + specific layer selection per ROI + Lasso CV model & 0.5552126635		 \\\hline
 resnet18 semi-supervised + specific layer selection per ROI + Lasso CV model & 0.5673601542 \\\hline
 
 resnet18 semi weakly supervised + specific 2 layers selection per ROI + RES model & 0.5113630713		 \\\hline

  resnet18 semi weakly supervised + higher dimension features + RES model & 0.5052717458		 \\\hline
 
resnet50 semi-supervised + specific layer selection per ROI + Lasso CV model & 0.5637263751 \\\hline
Squeezenet + specific layer selection per ROI + Lasso CV model & 0.4994109545 \\\hline
VOneNet resnset50 + specific layer selection per ROI + Lasso CV model & 0.5307863683 \\\hline
VOneNet Alexnet + specific layer selection per ROI + Lasso CV model & 0.4756440395 \\ \hline \hline
 
\end{tabular}
\caption{Results for mini-track (9 ROIs), scoring is based on Pearson’s correlation, resnet-18 semi-supervised outperform all other model in the table based on correlation score}
\label{results_minitrack}

\end{center}
\end{table}

\begin{table}[h]
\begin{center}
\begin{tabular}{||c | c||} 
 \hline
 Method & Score\\ [0.5ex] 
 \hline\hline
 Alexnet + all layers as features + Sparse linear regression & 0.1229460643  \\ 
 \hline
 AlexNet + specific layer selection per ROI + Sparse linear regression & 0.1893032764  \\
 \hline
 resnet18 + specific layer selection per ROI + Sparse linear regression & 0.196156364  \\
 \hline
 
 Nasenet mobile + no PCA + specific layer selection per ROI + RES model & 0.2640751352	 \\\hline
 VGG19 + specific layer selection per ROI +RES & 0.0149241953		 \\\hline
 resnet18 semi weakly supervised + specific 2 layers selection per ROI + RES & 0.2503818611		 \\\hline
 
 resnet18 semi-supervised + specific 2 layers selection per ROI + RES model & 0.2496098364 \\\hline
 
resnet50 semi-supervised + specific layer selection per ROI + RES model & 0.2613774908 \\\hline

VOneNet resnset50 + specific layer selection per ROI + RES model & 0.2491422229 \\\hline \hline

\end{tabular}
\caption{Results for the full-track (Whole-Brain), scoring is based on Pearson’s correlation, resnet-18 semi weakly-supervised outperform all other models in the table based on correlation score}
\label{results_fulltrack}

\end{center}
\end{table}

\section{Conclusion}
In this paper, the results and the main findings of the Algonauts Project 2021 Challenge: How the Human Brain Makes Sense of a World in Motion were presented, in which different experiments were carried out using two approaches: voxel-wise encoding model, and ROI-wise encoding models. In voxel-wise encoding methods semi-weakly supervised models and semi-supervised models with regularized regression seem to achieve high performance, while in ROI-wise encoding models, the model seems to be disturbed by noise in the voxel readings. Future work will include investigating more models, layers, and better validation techniques for voxel-wise encoding models, while different loss functions and regularization will be investigated for ROI-wise encoding models.

\bibliographystyle{unsrtnat}
\bibliography{references} 

\end{document}